\documentclass[9pt,twocolumn]{extarticle}
\usepackage[T1]{fontenc}
\usepackage[english,polish]{babel}
\usepackage[utf8]{inputenc}
\usepackage{authblk}
\usepackage{subfigure}
\usepackage{xcolor}
\usepackage{arydshln}
\usepackage{lmodern}
\usepackage{graphicx}
\usepackage{caption} 
\usepackage{mathtools}
\usepackage{amsfonts}
\usepackage{url}

\usepackage[a4paper,margin=2cm,footskip=1cm]{geometry}

\title{\textbf{An efficient manifold density estimator for all recommendation systems}}

\usepackage{authblk}
\author[1]{Jacek Dąbrowski}
\author[1,2]{Barbara Rychalska}
\author[1]{Michał Daniluk}
\author[1,2]{Dominika Basaj}
\author[1]{Konrad~Gołuchowski}
\author[1]{Piotr~Bąbel}
\author[1]{Andrzej~Michałowski}
\author[1]{Adam~Jakubowski}

\affil[1]{Synerise}
\affil[2]{Warsaw University of Technology}
\affil[ ]{\{jack.dabrowski, barbara.rychalska, michal.daniluk, dominika.basaj, konrad.goluchowski, piotr.babel, andrzej.michalowski, adam.jakubowski\}@synerise.com}
\date{}

\begin{document}
\maketitle
\selectlanguage{english}
\begin{abstract}
Many unsupervised representation learning methods belong to the class of similarity learning models. While various modality-specific approaches exist for different types of data, a core property of many methods is that representations of similar inputs are close under some similarity function. We propose EMDE (Efficient Manifold Density Estimator) - a framework utilizing arbitrary vector representations with the property of local similarity to succinctly represent smooth probability densities on Riemannian manifolds. Our approximate representation has the desirable properties of being fixed-size and having simple additive compositionality, thus being especially amenable to treatment with neural networks - both as input and output format, producing efficient conditional estimators. We generalize and reformulate the problem of multi-modal recommendations as conditional, weighted density estimation on manifolds. Our approach allows for trivial inclusion of multiple interaction types, modalities of data as well as interaction strengths for any recommendation setting. Applying EMDE to both top-k and session-based recommendation settings, we establish new state-of-the-art results on multiple open datasets in both uni-modal and multi-modal settings.
\end{abstract}
 
\section{Introduction}
The goal of recommender systems is to suggest items which a user might find interesting, often in the setting of online stores or social media. A common problem setting in the domain of recommenders is that predictions must be made from data which is inherently sequential, representing user actions over a time-span \cite{ludewig2018evaluation,ben2015recsys,nowplaying}. Thus, the input is an ordered collection of items based on a single user's shopping session, and the task consists of predicting which item will be clicked or added to cart next. Many session-based recommendation (SRS) systems use methods which explicitly model sequentiality, such as recurrent neural networks (RNNs) \cite{10.1145/3269206.3271761,10.1145/3132847.3132926,10.1145/3125486.3125491,10.1145/3397271.3401273} or graph neural networks (GNNs) \cite{Wu:2019ke,10.1145/3397271.3401319,10.1145/3394486.3403170}. However, these methods are known to scale poorly to growing item sets and increasing sequence size \cite{tallec2017unbiasing}. They also exhibit a number of specific efficiency-related problems, such as \textit{neighborhood explosion} in GNNs (the number of neighbors often grows exponentially when increasing node distances are considered). Such problems demand additional remedial measures which often hurt performance \cite{ZENG2021166, pmlr-v80-chen18p,Bai2020RippleWT}. Yet, as the sequential aspect of recommendation is considered vital, most efforts are focused on researching ever more complex neural network architectures in order to represent the ordered relations accurately. 
%On the other hand, the Transformer architecture \todo{cite} proves that inherently sequential data does not have to be processed in an ordered fashion to achieve state-of-the-art results in many machine learning areas.

\begin{figure}%
\centering
\includegraphics[width=80mm]{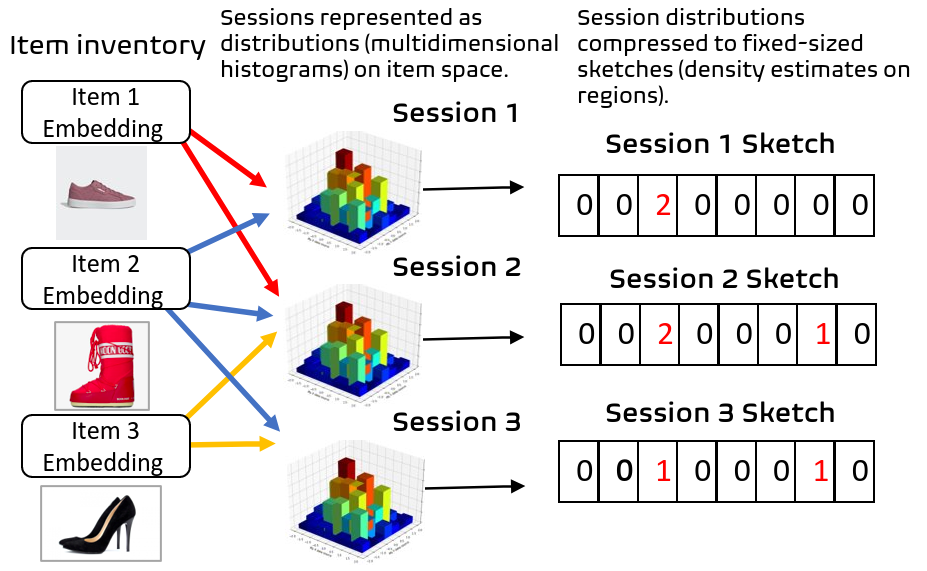}
\caption{Encoding densities in a multidimensional space. Each item is represented by an embedding vector of arbitrary modality (interactions, textual description, image). Embedding spaces of items are represented with multidimensional histograms. Items from each session are encoded into fixed-size \textit{sketch} structures.}
\label{fig:highlevel}
\end{figure}

We propose to shift the focus from sequentiality in recommenders to an accurate representation of user behaviors understood as item sets. Our hypothesis is that the sequential aspect could be largely discarded if we manage to represent the density of aggregate user preference spaces with sufficiently high fidelity. This approach could simplify the neural architecture dramatically, as probability densities do not have a temporal nature and can be represented by approximate, fixed-size structures, serving as simple 1-d input vectors to feed-forward neural networks.

Thus, the core challenge in our setting becomes to 1) create a succinct vector representation of densities, and 2) teach a model to compute conditional mappings between input and output densities. To this end, we propose EMDE (Efficient Manifold Density Estimator) -- a method which exploits locality sensitive hashing to create \textit{sketches} - histogram-like structures which represent density on multidimensional manifolds. The \textit{sketches} allow conditional mapping from one density estimate to another with simple shallow feed-forward neural networks. We show that EMDE achieves state-of-the-art results on multiple SRS datasets, going against the trend of sequential focus. We confirm the versatility of EMDE applying it to another recommender setting - top-k recommendation task, which operates on large item collections viewed/bought over an extended time period, and achieving competitive results. Thus, we postulate that direct density estimation of item spaces is indeed a viable research pathway in the general area of recommendation.

EMDE also exhibits a number of unique properties crucial to real-life recommendation tasks, which marks its efficiency against other neural methods: 
\begin{itemize}
    \item The \textit{sketches} have constant size independent of the total number of items modeled, the number of items a given user has interacted with, and the original embedding dimensions. Thus the size of a downstream model does not depend on these values as well.
    \item The \textit{sketches} are additive: aggregation of multiple samples to create item sets is done with a simple summation, with the option to include attached weights.
    % \item A simple concatenation of \textit{sketches} representing separate modalities is in practice good enough to grant high performance, simplifying the problem of multimodal fusion. 
    \item Retrieving items from a \textit{sketch} is simple and efficient (small, constant number of lookups per item).
\end{itemize}

 Moreover, we show that EMDE allows for easy and natural incorporation of multiple modalities, such as various types of interactions (click/purchase/add to favorites), item names or images. Multiple modalities can be introduced simply as additional manifolds whose density is estimated and mapped by EMDE to the same \textit{sketch} structures. A simple concatenation of per-modality \textit{sketches} is then fed to a shallow, feed-forward network. This feature is important even in a unimodal setting, because multiple versions of the same embedding can be easily used (e.g. computed with a different seed or iteration number) - a feature we exploit in our experiments.
 
 Our conclusions are backed by experiments  inspired by findings from  \cite{ludewig2019performance} and \cite{10.1145/3298689.3347058}. They observe the \textit{phantom progress} problem in recommender systems: carefully tuned simple heuristics (such as nearest-neighbor methods) in practice often outperform complex deep learning models, while algorithm performance is heavily dependent on the dataset and chosen performance metrics. In response to this, we use the benchmark suites from \cite{ludewig2019performance} and \cite{10.1145/3298689.3347058} to test our approach on a wide range of referential metrics and datasets, outperforming both the most advanced models and simple heuristics alike.

\begin{figure*}[ht]%
\centering
\includegraphics[width=150mm]{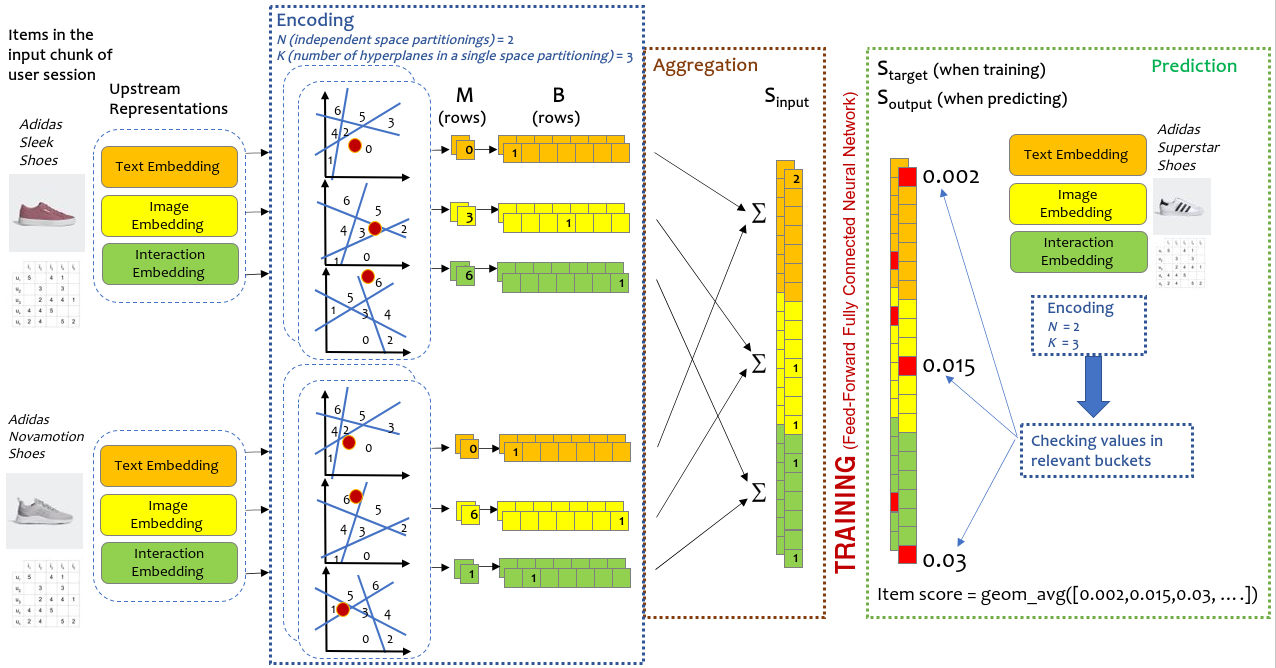}
\caption{EMDE recommender architecture.}
\label{fig:architectur}
\end{figure*}

\section{Related Work}

\begin{figure}%
\centering
\includegraphics[width=60mm]{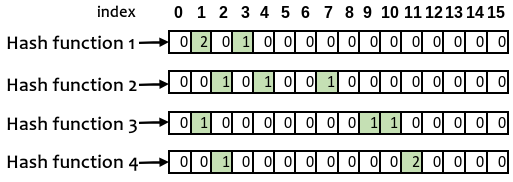}
\caption{An example of sketch produced by the count-min sketch algorithm. The displayed sketch stores three items, with hash collisions of two items in buckets containing the value 2.}
\label{fig:cms}
\end{figure}

\textbf{Neural session-based recommenders.}
A popular branch of neural models use recurrent neural networks (RNN), suitable for data with temporal ordering. Gru4Rec \cite{10.1145/3269206.3271761} uses the gated recurrent unit (GRU), in a sequential setting. NARM \cite{10.1145/3132847.3132926} extends the approach with an item-level attention mechanism. STAMP \cite{10.1145/3219819.3219950} is an attention-based non-RNN model using neural memory modules to represent item long-term and short-term sequences. NextItNet \cite{DBLP:journals/corr/abs-1808-05163} is based on stacked 1D dilated convolutions, instead of an RNN. SR-GNN \cite{Wu:2019ke} features a graph neural network (GNN) able to exploit additional paths between sessions, still focusing on sequentiality. TAGNN \cite{10.1145/3397271.3401319} adds a target-aware attention module to the GNN. The GNN-based models do not scale easily and are usually evaluated on small datasets (e.g. \cite{10.1145/3397271.3401319} evaluate on just 1/64 of the RSC15 dataset, while EMDE and most methods evaluated by \cite{ludewig2019performance} work on the full dataset). 

% Our work differs from these approaches in that it represents sequentiality in only by weighting items while adding them into item set, and uses a simple and lightweight feed-forward network.

\textbf{Non-neural session-based recommenders.} Although neural recommenders are mostly regarded as state-of-the art, \cite{ludewig2019performance} show that simple non-neural baselines can often outperform them in practice. They propose S-KNN - a session-based nearest-neighbor technique, and its variant VS-KNN which favors recent items in sessions. AR is based on association rules, with its variant SR counting associations in item sequences. CT \cite{Mi2018ContextTF} uses context trees for reflecting possible decision pathways during sessions. %\cite{ludewig2019performance} show that these approaches cannot be disregarded and we follow, evaluating against both simple and complex competitors.

\textbf{Sketching-based density estimators.} \cite{charikar2018hashingbasedestimators} introduce methods for kernel density estimation (KDE) based on locality-sensitive hashing (LSH). Subsequently \cite{siminelakis2019rehashing} use a sketch-based structure for a compressed representation of multiple LSH partitions for KDE. Both methods require a computation-heavy sampling procedure to arrive at density estimates. \cite{coleman2019sublinear} introduce RACE - a LSH sketch-based method for KDE, which does not require sampling to arrive at density estimates. \cite{coleman2019race} further explores the technique of LSH sketching for approximate nearest-neighbor search on streaming data. In these methods, the considered manifold is $R^n$.

\section{Algorithm}

\subsection{Preliminaries}
Our solution is inspired by two  algorithms: count-min sketch (CMS) \cite{CMS} and locality sensitive hashing (LSH) \cite{lsh_indyk}. 

\textbf{Count-min sketch.} CMS is a data structure used to count the number of times items appear in a data stream. The algorithm provides efficiency at the cost of accuracy: it can encode multi-sets in an compact, compressed way, but returned item counts are approximate. Often used when the input space is too large to fit in memory, the algorithm works in sublinear space (Figure \ref{fig:cms}). It consists of two operations: 1) incrementing item counts; 2) querying item counts by computing a minimum across hashed tables. The CMS \textit{sketch} structure is represented by a two-dimensional array $count$ with width $w$ and depth $d$, representing hash functions $h_d$, which output a value $h_d (i_{t}) \in (0,..,w-1)$. Each row of the array represents a separate frequency table, and multiple such tables are introduced in order to alleviate the problem of hash conflicts. The array $count_{d \times w}$ is initialized with zeros. When an item $i_t$ arrives, its count is updated by a value of $c_t$ (1 by default). The procedure is expressed by the following equation: \[count_d[h_d (i_t)] \leftarrow count_d[h_d(i_t)]+c_t.\]

The retrieval procedure of an item count consists of checking the values contained in the $d$ respective item buckets in each row of the sketch. The minimum value of all of these is finally selected:
\[item\_count = \min_{j=1}^{d} count_{j}[h_j(i_t)].\]

While CMS is a very efficient, compressed data structure for count estimation, the algorithm applies random hashes to item labels, therefore is blind to the geometry of the input space and any item-item relationships.

\textbf{Locality-sensitive hashing.} As a geometry-aware approach to hashing, LSH aims to assign input vectors to hash codes, ensuring that the probability of being in the same hash bucket is much higher for inputs which are close together in the input space, than for those which are far apart \cite{lsh_indyk}.

In contrast to CMS random hashing, LSH methods allow to preserve the geometric prior of the original input space when available. As we further show, CMS and LSH can be combined and adapted for compressed, geometry-aware density estimation in high dimensional spaces.

\subsection{Efficient Manifold Density Estimator}\label{EMDE}

EMDE can be likened to a weighted, compressed, geometry-aware histogram for high dimensional spaces. Similarly to a histogram accumulating normalized observation counts in disjoint regions of a 1-dimensional input space, EMDE also functions as a piece-wise constant observation counter over disjoint regions of an input space lying on the data manifold. As dimensionality of the input space grows, performance of a simple histogram degrades dramatically. Thus, to efficiently map high dimensional spaces to a histogram-like structure, we utilize the CMS compression approach - maintaining an ensemble of multiple, independent histograms. Originally CMS uses random hash functions to map inputs onto sketch buckets, which is unsuitable for continuous spaces. Instead, we utilize data-dependent LSH methods, to ensure a piecewise-continuous mapping, analogous to histogram buckets. The resulting partitionings define regions of the data manifold corresponding to CMS input-to-bucket mapping. EMDE operates on manifolds spanned by vector representations coming from upstream metric representation learning methods (text/image/graph node embeddings). This helps to ensure that semantically similar items frequently share their assigned buckets in the sketch, preserving the geometric prior from upstream embeddings. Combining the compressive properties of CMS, the LSH inspired preservation of geometric priors, with the simplicity of histograms, EMDE is a form of a probability density estimator which can efficiently scale to extremely large input dimensionalities.

A typical application of EMDE consists of 4 steps: 1) computing multiple independent partitionings of the data manifold (analogous to computing the boundaries of histogram buckets), usually done once; 2) filling the resulting structure with weighted observations, done once for each set of observations (e.g. single user's interactions) to be summarized; 3) using a simple neural network to model input density into output density transformation 4) querying the output structure to obtain density estimates. Step 3) is optional and required only for conditional estimation, it can be skipped in case of simple density estimation.

\subsection{Session-based Recommendation with EMDE}

Below we describe EMDE on the example of item recommendation. The procedure is also displayed in Figure \ref{fig:architectur}. We focus on the scenario of SRS, where the data is comprised of user interaction sessions. Based on the items which have appeared in the session so far (\textit{input chunk}), our objective is to predict the item which will be purchased/clicked next. The application of EMDE to SRS proceeds in the following steps:

\subsubsection{Obtaining upstream item embeddings} As the first step we need to precompute input embeddings using an external embedding method. EMDE can handle all sorts of embeddings, for example capturing interaction data (often interactions of different types, e.g. click, purchase, favorite), item names, attributes, or images. In practice, virtually all popular embedding schemes fulfill the theoretical conditions which make them applicable for our method (see Section \ref{sec:local-similarity} for details).

\subsubsection{Encoding} 
A central concept of EMDE is the data manifold $\mathbb{M}$
%:= h_{\theta}(\mathbb{X})$ 
locally embedded in $\mathbb{R}^{n}$.
The manifold $\mathbb{M}$ is spanned by embedding vectors computed for a particular data modality. Our objective is to perform multiple partitionings of of the data manifold into regions capturing the local metric prior induced by the underlying embedding vectors (see \textit{Encoding} section in Figure~\ref{fig:architectur}). As such, semantically similar inputs should be assigned to the same manifold regions frequently. 
% Our objective is to perform multiple independent partitionings of the manifold, into $2^K$ local regions each.
In a statistical interpretation, this operation corresponds to computing bucket boundaries of histograms, given samples from their input distributions.
% This operation is done once for a given data modality and allows subsequent aggregation of different item sets.

To this end, we propose a modified version of LSH algorithm we call Density-dependent LSH (DLSH). We start with choosing $K$ random vectors $r_i$, then for $v \in \mathbb{M}$ we let $hash_{i}(v) = sgn(v\boldsymbol{\cdot}r_{i} - b_{i})$, where the bias value $b_{i}$ is drawn from $Q_{i}(U \sim \operatorname{Unif}[0,1])$
%Q_{i}(0.5)$
, where $Q_{i}$ is the quantile function of $\{v\boldsymbol{\cdot}r_{i} : v \in \mathbb{M}\}$. In contrast to LSH, this scheme is density-dependent, cutting the the manifold into non-empty parts, thus avoiding unutilized regions.
%equally dense halves, and maximizing the entropy of the partitioning. 
As each hash independently divides the manifold into two parts, we treat these $K$ binary hashcodes as bits of a $K$-bit integer. We thus obtain a partitioning of the manifold into $2^K$ disjoint regions (or $\sum_{i=0}^n \binom{K}{i}$ if $K > n$; we further assume $K < n$ for simplicity), which correspond to the set of all geometric intersections of regions spanned by single-bit hashes. The $2^K$ regions of the manifold form a single \textit{depth} level of our sketch, analogous to a simple histogram with $2^K$ buckets, but in multiple dimensions. As a piecewise-constant estimator, the resolution of a histogram is quite poor - being only capable of outputting as many different values, as there are buckets / regions, with expressivity scaling linearly with structure size.

To overcome linear scaling with respect to the number of regions, we use a procedure inspired by CMS. Instead of increasing the number of regions into which a manifold is partitioned, we maintain $N$ independent partitionings, akin to different \textit{depth} levels of a CMS. Since any given point is located in a single region (out of $2^K$) of each independent partitioning (out of $N$), we know that it also lies in the geometric intersection of all $N$ regions it belongs to. This intersection can be extremely small, which leads to an exponential growth of the structure's resolution power.

Thus we perform the above partitioning procedure $N$ times independently, starting with new random vectors $r_i$ every time, resulting in a \textit{sketch structure} of \textit{width} $2^K$ and \textit{depth} $N$, named by analogy to CMS.
In Figure \ref{fig:architectur} item embeddings are encoded into a sketch of depth $N=2$ and $K=3$.
% Local similarity of the data manifold allows DLSH to assign semantically similar inputs to the same sketch regions frequently. This effect captures the local metric prior induced by the underlying representation learning method which produced the vectors $\mathbb{M}$.

We apply DLSH to obtain separate partitionings for each input modality. In item recommendation context, for each item, we store \textit{N} region indices (or \textit{buckets}) in per-modality matrices $\boldsymbol{M}_{n\_items \times N}$. The region indices stored in $\boldsymbol{M}$ are integers in the range $[0, 2^K-1]$ and form a sparse item encoding, suitable for efficient storage. In Figure \ref{fig:architectur}, for item \textit{Adidas Sleek Shoes}, its row in M is equal to $[0, 3, 6]$.

For the purposes of item-set aggregation, we one-hot encode each region index from the per-modality matrices $\boldsymbol{M}$, obtaining per-modality matrices $\boldsymbol{B}_{n\_items \times j}$, where $j\in \{1,\dots,{N} \times 2^{K}\}$. 
%For the purposes of item-set aggregation, we convert sparse item representation into dense per-modality matrices $\boldsymbol{B}_{i \times j}$, where $i\in \{1,\dots,n\_items\}$ and $j\in \{1,\dots,{N} \times 2^{K}\}$. 
Each row of the matrix $\boldsymbol{B}$ represents an item \textit{sketch}. Such an item sketch can be interpreted as $N$ concatenated histograms of $2^K$ buckets each. 
%Each such histogram contains a single value of $1$ inserted into the bucket assigned to the item's embedding vector.

% In all further steps, we hold the \textit{sketch structure} (partitioning of manifolds into regions) fixed.

\subsubsection{Aggregation}

Representations of (weighted) item multi-sets are obtained by simple elementwise (weighted) summation of individual item sketches. This follows from the additive compositionality of histograms with the same bucket ranges, or alternatively from the definition of CMS. Thanks to this property, sketches can also be constructed incrementally from data streams, which simplifies their application in industrial settings. Any subsequent normalization of sketches is performed along the \textit{width} dimension, as every level of \textit{depth} represents a separate, independent histogram.

% An empty $sketch$ is instantiated as 2-dimensional array $N \times 2^K$ and can be indexed by the outputs of $V$. Filled with zeros, it represents a degenerate zero density. We add samples $x \in H$ weighted by $w \in R^{+}$ from a smooth probability measure on the manifold $\mathbb{M}$, where $f$ is the probability density function,
% % satisfying $Pr(A)=\int_{A} f d \mu$
% by performing $sketch_{samples}[V(x)] := sketch_{samples}[V(x)] + w_x$ for all $x \in H$ and their corresponding weights $w \in R^{+}$. Our final representation is $sketch := sketch_{samples} / \sum{w_{x}}$.
% In the recommendation setting, aggregate multi-item sketches serve as representations of users' historic and future interactions during sessions. For a given user and some temporal split into past and future chunks of a session, we instantiate two separate empty sketch structures, as a 2-dimensional zero-filled arrays $N \times 2^K$, for both input (historic) and target (future) interactions. For each item in the input and target, we increment respectively the input or target sketch, by elementwise summation with the item's sketch (optionally multiplied by a weight). 

For a given user and some temporal split into past and future chunks of a session, we encode each input and target item into dense sketches representation (in form of matrix $\boldsymbol{B}$) and then we aggregate sketches of input items into single sketch $\boldsymbol{S_{input}}$ and sketches of target items into target sketch $\boldsymbol{S_{target}}$. The aggregations are performed separately for each modality. While in our setting, the target sketch contains only a single item to be predicted, the general formulation admits multiple target items. 
%We repeat this procedure for each data modality and concatenate the results, obtaining a pair of aggregate sketches $\boldsymbol{S_{input}}$ and $\boldsymbol{S_{target}}$ for each user session.
We arrive at a concise vector representation of weighted, multi-modal multi-sets, efficiently representing user interaction behavior.

\subsubsection{Model and loss function}
We use a simple feed-forward neural network as a conditional density estimator with sketch structures for all inputs, outputs and targets - making both the model size and training time fully independent of the total number of items, original upstream embedding sizes, or the lengths of user sessions.

As a simple $L_1$-normalized histogram can be considered to approximate a probability mass function of a discrete distribution, sketches normalized across \textit{width} are ensembles of such histograms for many individual distributions. We train our model to minimize mean Kullback–Leibler divergence between individual target and output probability mass functions in the ensemble. For every row in a batch, this entails: 1) $L_1$-normalizing the target sketch $\boldsymbol{S_{target}}$ across \textit{width}; 2) applying the $Softmax$ function to the network output logits $\boldsymbol{S_{output}}$ across \textit{width}; 3) calculating \textit{KL-divergence} between targets and outputs across \textit{width}; 4) averaging the loss across \textit{depth}. It is worth noting that our formulation bypasses the need for either: a) an extremely wide softmax layer with weight vectors for all items; b) negative sampling; both of which negatively affect stability and training time of standard methods.

To improve the stability of pre-activations in the neural network, we $L_2$-normalize the input \textit{sketch} $\boldsymbol{S_{input}}$ across \textit{width}.

\subsubsection{Prediction}
In order to produce a recommendation score for an item, we query the output sketch in a similar way to CMS queries. As the matrix $\boldsymbol{M}$ contains mappings of every item, to a single histogram bucket for all levels of sketch \textit{depth}, a simple lookup of $\boldsymbol{S_{output}}$ at the respective indices from $\boldsymbol{M}$ is sufficient. This operation can also be efficiently realized in dense batch format amenable to GPU acceleration, as matrix multiplication $\boldsymbol{S_{output}} \times \boldsymbol{B}$. Both versions yield $N$ independent probability estimates from individual elements of the depth-wise ensemble for each item. We aggregate the estimates using \textit{geometric mean} (see \textit{Prediction} section in Figure \ref{fig:architectur}). The difference between CMS, which uses \textit{minimum} operation for aggregation, and our sketches stems from operating on counts (CMS) versus probabilities (EMDE). Strong theoretical arguments for optimality of \textit{geometric mean} can be found in \cite{dognin2019wasserstein} and \cite{itoh2017geometric}, given the interpretation of sketches as an ensemble of density estimators over manifolds. We verify the choice of aggregation function empirically in Table \ref{ablation results}. The resulting probability estimates are unnormalized i.e. don't sum to 1 across all items (normalization can be performed but is unnecessary for the recommendation setting).

% \end{enumerate}

\subsection{EMDE Input: the Requirement of Local Similarity}\label{sec:local-similarity} EMDE uses DLSH to preserve the metric prior from its input space. Intuitively, the transferred prior has 3 effects: 1) a local smoothing effect for estimates on the data manifold; 2) errors due to hash collisions have a high likelihood to mistake an item for a  semantically similar one (still relevant for recommendation); 3) in neural models a hash bucket holding semantically similar items helps to reduce variance and speed up convergence, as opposed to a hash bucket with random items.

In the absence of any local geometry (e.g. fully random item embeddings), EMDE degenerates to a normalized CMS with random hash codes. In Section \ref{ablations} we show that such a geometry-blind approach yields sub-par results. Thus, it is essential for our method that the property of \textit{local similarity} under a locally-Euclidean metric holds at least approximately in the input space.

% The goal of deep metric representation learning is to learn a function $h_{\theta}(x): {\mathbb{X}} \rightarrow \mathbb{R}^{n}$ mapping inputs from the data manifold in $\mathbb{X}$ onto points in $\mathbb{R}^{n}$ which are metrically close if and only if they are semantically similar. In practice, $h_{\theta}(\mathbb{X}) \subset \mathbb{R}^{n}$ forms a Riemannian manifold locally embedded in Euclidean space.

Naively, it would seem that not all popular methods of deep representation learning follow the metric learning paradigm, optimizing e.g. skip-gram, masked-language-model, next sentence prediction, CPC, InfoNCE or DeepInfoMax objectives \cite{mikolov2013distributed, devlin-etal-2019-bert, oord2018representation, hjelm2018learning}. Fortunately, \cite{kong2019mutual} show that all the aforementioned self-supervised objectives correspond to InfoNCE, while \cite{tschannen2019mutual} observe that InfoNCE has a direct formulation in terms of metric learning. Thus, most existing representation learning methods exhibit local similarity and can be utilized by our method to capture the metric prior.

\begin{table*}[t]
  \caption{Session-based recommendation results. Top 5 competitors are displayed for each dataset. * are from \cite{10.1145/3298689.3347041}}
  \label{session-rec-results}
  \centering
  \small
  \setlength\tabcolsep{5pt}
  \scalebox{0.8}{
  \begin{tabular}{llllll||llllll}
  \hline
    %\toprule
    Model & MRR@20 & P@20 & R@20 & HR@20 & MAP@20 & Model & MRR@20 & P@20 & R@20 & HR@20 & MAP@20 \\\hline
        \multicolumn{6}{c||}{RETAIL} & \multicolumn{6}{c}{RSC15}\\\hline
    
    \textsc{EMDE} & \textbf{0.3524}  & 0.0526 & \textbf{0.4709} & \textbf{0.5879} & 0.0282 &
    \textsc{EMDE} & \textbf{0.3104} & 0.0730 & 0.4936 & 0.6619 &	0.0346
    \\ \hdashline
    
    \textsc{VS-KNN*} & 0.3395  &	\textbf{0.0531} &	0.4632 &	0.5745 & 0.0278 &
    \textsc{CT \cite{DBLP:journals/corr/abs-1806-03733}} & 0.3072  &	0.0654 &	0.471 &	0.6359 & 0.0316
    \\
    
    \textsc{S-KNN*} & 0.337  &	0.0532 &	0.4707 &	0.5788 &	\textbf{0.0283} &
    \textsc{NARM \cite{10.1145/3132847.3132926}} & 0.3047  &	\textbf{0.0735} &	\textbf{0.5109} &	\textbf{0.6751} & \textbf{0.0357}
    \\
    
    \textsc{TAGNN \cite{10.1145/3397271.3401319}} & 0.3266 & 0.0463 & 0.4237 & 0.5240 & 0.0249 &
    \textsc{STAMP \cite{10.1145/3132847.3132926}} & 0.3033 & 0.0713  &	0.4979 & 0.6654 & 0.0344
    \\
    
    \textsc{Gru4Rec \cite{hidasi2015sessionbased}} & 0.3237  &	0.0502 & 0.4559 & 0.5669 & 0.0272 &

    \textsc{SR \cite{Kamehkhosh2017ACO}} & 0.301  &	0.0684 &	0.4853 &	0.6506 & 0.0332 
    \\
    
        \textsc{NARM \cite{10.1145/3132847.3132926}} & 0.3196 & 0.044	& 0.4072 & 0.5549 & 0.0239  &

    \textsc{AR \cite{asso_rules}} & 0.2894  &	0.0673 &	0.476 &	0.6361 & 0.0325
    \\
    \hline
    \multicolumn{6}{c||}{DIGI} & \multicolumn{6}{c}{NOWP} \\\hline

    \textsc{EMDE} & 0.1724 & \textbf{0.0602} &	\textbf{0.3753} & \textbf{0.4761} &	\textbf{0.0258} &
    \textsc{EMDE} & \textbf{0.1108}  & \textbf{0.0617} & \textbf{0.1847} & \textbf{0.2665} & 0.0179 \\ \hdashline
    
    \textsc{VS-KNN*} & \textbf{0.1784} &	0.0584 &	0.3668 &	0.4729 &	0.0249 &
    \textsc{CT \cite{DBLP:journals/corr/abs-1806-03733}} & 0.1094 &	0.0287 &	0.0893 &	0.1679 &	0.0065
    \\
    \textsc{S-KNN*} & 0.1714 &	0.0596 &	0.3715 & 0.4748 &	0.0255 & 
    \textsc{Gru4Rec \cite{hidasi2015sessionbased}} & 0.1076 &	0.0449 &	0.1361 &	0.2261 &	0.0116
    \\
    
    \textsc{TAGNN} \cite{10.1145/3397271.3401319} & 0.1697 & 0.0573 & 0.3554 & 0.4583 & 0.0249 &
    \textsc{SR \cite{Kamehkhosh2017ACO}} & 0.1052  &	0.0466 &	0.1366 &	0.2002 &	0.0133
    \\
    
        \textsc{Gru4Rec \cite{hidasi2015sessionbased}} & 0.1644 &	0.0577 &	0.3617 &	0.4639 &	0.0247 &

    \textsc{SR-GNN \cite{DBLP:journals/corr/abs-1811-00855}} & 0.0981 &	0.0414 &	0.1194 &	0.1968 &	0.0101
    \\
    
        \textsc{SR-GNN \cite{DBLP:journals/corr/abs-1811-00855}} &	0.1551 &	0.0571 &	0.3535 &	0.4523 &	0.0240 & 

    \textsc{S-KNN*} & 0.0687	& 0.0655 &	0.1809 &	0.245 &	\textbf{0.0186} \\
    
    \hline
        \multicolumn{6}{c||}{30MU} &  \multicolumn{6}{c}{AOTM} \\\hline
    \textsc{EMDE} & \textbf{0.2673}	& \textbf{0.1102} & \textbf{0.2502} & \textbf{0.4104} & \textbf{0.0330} & 
    \textsc{EMDE} & \textbf{0.0123}  & 0.0083 & 0.0227 & 0.0292 & 0.0020 \\ \hdashline
    
    \textsc{CT \cite{DBLP:journals/corr/abs-1806-03733}} & 0.2502 &	0.0308 &	0.0885 &	0.2882 &	0.0058 &
    \textsc{CT \cite{DBLP:journals/corr/abs-1806-03733}} & 0.0111 &	0.0043 &	0.0126 &	0.0191 &	0.0006 \\
    
    \textsc{SR \cite{Kamehkhosh2017ACO}} & 0.241 &	0.0816 &	0.1937 &	0.3327 &	0.024 &
    \textsc{NARM \cite{10.1145/3132847.3132926}} & 0.0088 &	0.005 &	0.0146 &	0.0202 &	0.0009 \\
    
    \textsc{Gru4Rec \cite{hidasi2015sessionbased}} & 0.2369 &	0.0617 &	0.1529 &	0.3273 &	0.015 & 
    \textsc{STAMP \cite{10.1145/3132847.3132926}} & 0.0088 &	0.002 &	0.0063 &	0.0128 &	0.0003 \\
    
    \textsc{NARM \cite{10.1145/3132847.3132926}} & 0.1945 &	0.0675 &	0.1486 &	0.2956 &	0.0155 & 
    \textsc{SR \cite{Kamehkhosh2017ACO}} & 0.0074 &	0.0047 &	0.0134 &	0.0186 &	0.001 \\
    
    \textsc{VS-KNN*} & 0.1162 &	0.109 &	0.2347 &	0.383 &	0.0309 &

    \textsc{S-KNN*} & 0.0054 &	\textbf{0.0139} &	\textbf{0.039} &	\textbf{0.0417} & \textbf{0.0037} \\
    
    \hline
    %\bottomrule
  \end{tabular}
  }
\end{table*}

\begin{table}[t]
  \caption{Results of adding multimodal data to our session-based EMDE recommenders.}
  \label{session-rec-mm-results}
  \centering
  \tiny
  \begin{tabular}{llllll}
  \hline
    %\toprule
    Model & MRR@20 & P@20 & R@20 & HR@20 & MAP@20  \\\hline
        \multicolumn{6}{c}{RETAIL} \\\hline
    \textsc{EMDE MM} & \textbf{0.3664} & \textbf{0.0571} & \textbf{0.5073} & \textbf{0.6330} & \textbf{0.0309} \\
    % \textsc{EMDE} & 0.3522 & 0.0512 & 0.4643 & 0.5774 & 0.0277 \\\hline
    
    \textsc{EMDE} & \textbf{0.3524}  & 0.0526 & \textbf{0.4709} & \textbf{0.5879} & 0.0282 \\\hline
    
    \multicolumn{6}{c}{RSC15}\\\hline         \textsc{EMDE MM} & \textbf{0.3116}  & \textbf{0.0743} & \textbf{0.5000} & \textbf{0.6680} & \textbf{0.0352} \\

    \textsc{EMDE} & 0.3104 & 0.0730 & 0.4936 & 0.6619 &	0.0346
    
    \\\hline
    \multicolumn{6}{c}{DIGI} \\\hline
    \textsc{EMDE MM} & \textbf{0.1731} & \textbf{0.0620} &\textbf{0.3849} &	\textbf{0.4908} &  \textbf{0.0268} \\
    
    \textsc{EMDE} & 0.1724 & \textbf{0.0602} &	\textbf{0.3753} & \textbf{0.4761} &	\textbf{0.0258} \\\hline
    
\multicolumn{6}{c}{30MU} \\\hline
\textsc{EMDE MM} & \textbf{0.2703} & \textbf{0.1106} & \textbf{0.2503} & \textbf{0.4105} & \textbf{0.0331}
    \\
\textsc{EMDE} & 0.2673	& 0.1102 & 0.2502 & 0.4104 & 0.0330
\\ \hline

    %\bottomrule
  \end{tabular}
\end{table}

\section{Experiments}

\subsection{EMDE as a Recommender System}
We report results for unimodal \texttt{EMDE} (no multimodal data, only basic user-item interactions found in the original datasets), and \texttt{EMDE MM} (configurations where selected multimodal channels are present). We evaluate all algorithms in the same frameworks of \cite{10.1145/3298689.3347058} and  \cite{ludewig2019performance}, keeping to their selected performance measures and datasets. In order to disentangle the gains induced by our method from the quality of the  embeddings themselves, we use a very simple graph node embedding scheme \cite{cleora}, which is the same for representing both text and interaction networks. For embedding text we simply create a graph of item-token edges. We leave experiments with elaborate embeddings such as BERT \cite{devlin-etal-2019-bert} for future research.

We conduct our experiments on a machine with 28 CPUs, 128 GB RAM and a commodity nVidia GeForce RTX 2080 Ti 11GB RAM GPU card.

\begin{table*}[t]
  \caption{Top-k recommendation results.}
  \label{top-k-results}
  \centering
  \setlength\tabcolsep{2.5pt}
  \small
  \begin{tabular}{l|lllllll}
  %\toprule
  \hline
  Model & Recall\newline@1 & NDCG\newline@5 & Recall\newline@5 & NDCG\newline@10 & Recall\newline@10 & NDCG\newline@20 & Recall\newline@20 \\ \hline
  \multicolumn{8}{c}{MovieLens 20M} \\\hline
  EMDE  & \textbf{0.0529} & \textbf{0.2017} & \textbf{0.1662} & \textbf{0.2535} & \textbf{0.2493} & 0.3053 & 0.3523 \\ \hdashline
  EASE \cite{10.1145/3308558.3313710} & 0.0507  & 0.1990 & 0.1616 & 0.2530 & 0.2465 & \textbf{0.3078} & \textbf{0.3542}  \\ 
  MultVAE \cite{10.1145/3178876.3186150} & 0.0514 & 0.1955 & 0.1627 & 0.2493 & 0.2488 & 0.3052 & 0.3589 \\
  SLIM \cite{10.1109/ICDM.2011.134} &0.0474 &0.1885 &0.1533 &0.2389 &0.2318 &0.2916 &0.3350 \\ 
  RP3beta \cite{rp3b} &0.0380 &0.1550 &0.1279 &0.2004 &0.2007 &0.2501 &0.3018 \\ 
 \hline
  \multicolumn{8}{c}{Netflix Prize} \\\hline
  EMDE & \textbf{0.0328} & 0.1512 & 0.1101 & 0.1876 & 0.1652 & 0.2332 & 0.2432  \\ \hdashline
  EASE \cite{10.1145/3308558.3313710} & 0.0323 & \textbf{0.1558} & \textbf{0.1120} & \textbf{0.2050} & \textbf{0.1782} & \textbf{0.2589} & \textbf{0.2677} \\
  MultVAE \cite{10.1145/3178876.3186150} & 0.0313 &  0.1485 & 0.1109 & 0.1957 & 0.1756 & 0.2483 & 0.2645 \\
  SLIM  \cite{10.1109/ICDM.2011.134}    &0.0307 &{0.1484}        &0.1062 &0.1952 &0.1688 &0.2474 &0.2552 \\

RP3beta \cite{rp3b} &0.0243 &0.0946 &0.0595 &0.1191 &0.0863 &0.1578 &0.1390  \\ 
  \hline

\hline
  \end{tabular}
\end{table*}

\begin{table}[t]
  \caption{Results of adding multimodal data to our top-k EMDE recommenders.}
  \label{tab:top-k-mm-results}
  \centering
  \setlength\tabcolsep{2.5pt}
  \tiny
  \begin{tabular}{p{0.8cm}|p{0.8cm}p{0.8cm}p{0.8cm}p{0.8cm}p{0.8cm}p{0.8cm}p{0.8cm}}
  %\toprule
  \hline
  Model & Recall@1 & NDCG@5 & Recall@5 & NDCG@10 & Recall@10 & NDCG@20 & Recall@20 \\ \hline
  \multicolumn{8}{c}{MovieLens 20M} \\\hline
  EMDE MM &  \textbf{0.0670} & \textbf{0.2378} & \textbf{0.1963} & \textbf{0.2890} & \textbf{0.2780} & \textbf{0.3358} & \textbf{0.3710} \\
  EMDE  & 0.0529 & 0.2017 & 0.1662 & 0.2535 & 0.2493 & 0.3053 & 0.3523 \\ 
 \hline
  \multicolumn{8}{c}{Netflix Prize} \\\hline
  EMDE MM & \textbf{0.0388} & \textbf{0.1574}  & \textbf{0.1253} & \textbf{0.2155} & \textbf{0.1875} & \textbf{0.2619} & \textbf{0.2645} \\
%   EMDE & 0.0310 & 0.1448 & 0.1059 & 0.1876 & 0.1652 & 0.2332 & 0.2432 \\
EMDE & 0.0328 & 0.1512 & 0.1101 & 0.1876 & 0.1652 & 0.2332 & 0.2432  \\
  \hline
  \end{tabular}
\end{table}

\begin{figure}
\centering

\subfigure[GloVe]{%
  \includegraphics[width=0.19\textwidth]{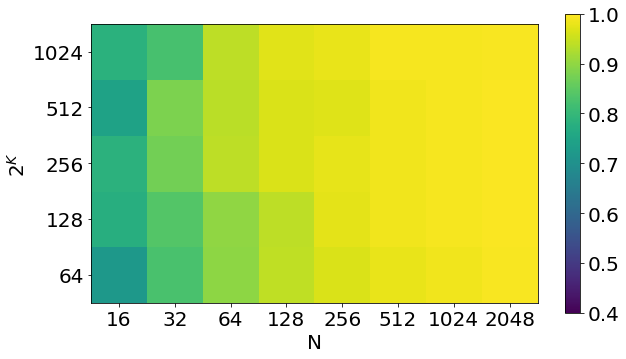}%
}
\subfigure[MNIST]{%
  \includegraphics[width=0.19\textwidth]{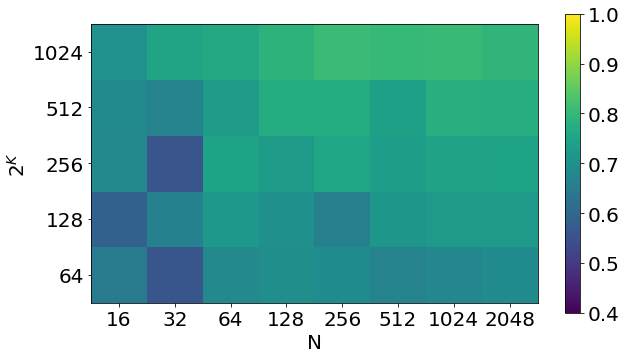}%
} \\
\subfigure[MovieLens20M]{%
  \includegraphics[width=0.19\textwidth]{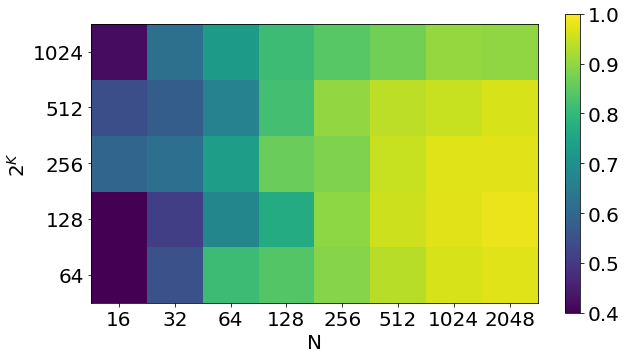}%
}
\subfigure[Netflix]{%
  \includegraphics[width=0.19\textwidth]{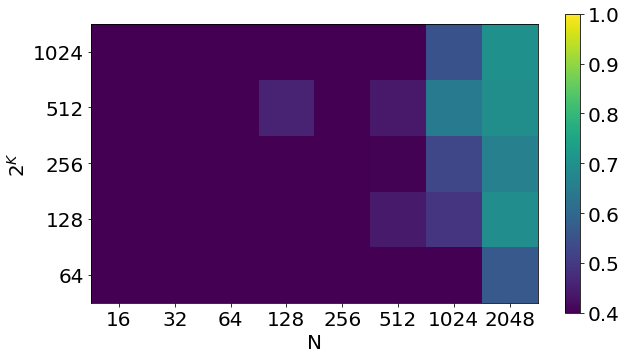}%
}
\caption{Density estimation quality with various N/K configurations. Metric reported is Pearson correlation coefficient against true distribution.}
\label{f:de_results}
\end{figure}

\subsubsection{Session-Based Recommendation}\label{session_experiment}

We conduct experiments on six popular datasets used in \cite{ludewig2019performance}.  We reuse data preprocessing, training framework, hyperparameter search procedure and evaluation from the framework of \cite{ludewig2019performance}\footnote{\url{https://github.com/rn5l/session-rec}} without introducing any changes. Within the framework, each dataset is split into five contiguous in time slices in order to minimize the risk of random effects. For each dataset we locate and use multimodal data whenever possible. We experiment on the following datasets:
\begin{itemize}
\item \textit{RETAIL} dataset has been collected from a real-world e-commerce website. It consists of behaviour data and item properties such as price, category, vendor, product type etc. The dataset consists of 212,182 actions in 59,962 sessions over 31,968 items.

% We perform small random search over properties and found that \textit{839}, \textit{776} and \textit{6} properties are most useful.

\item \textit{DIGI} is an e-commerce dataset shared by the company Diginetica, introduced at a CIKM 2016 challenge. The goal is to predict next item view in each session. As multimodal data we use purchases, product name tokens and items returned by search queries.

\item \textit{RSC15} also known as Yoochoose dataset, is an e-commerce dataset used in the 2015 ACM RecSys Challenge. It contains 5,426,961 actions in 1,375,128 sessions over 28,582 items. Purchased products were used as additional data source.

\item \textit{30Music} dataset consists of music listening logs obtained from Last.fm. It contains 638,933 actions in 37,333 sessions over 210,633 items. Playlists created by the users were used as additional multimodal data.

\item \textit{AOTM} is a music dataset containing user-contributed playlists from the Art of the Mix webpage. It contains 306,830 actions in 21,888 sessions over 91,166 items.

\item \textit{NOWP} contains song playlists collected from Twitter. It consists of 271,177 actions in 27,005 sessions over 75,169 items. 
\end{itemize}
For \textit{AOTM} and \textit{NOWP} datasets, we were unable to find multimodal data with matching item identifiers.

\textbf{Benchmarks.} We compare against the benchmark methods from \cite{ludewig2019performance}, which have been fine-tuned extensively by the benchmark's authors. We include two recent graph neural models: SR-GNN  \cite{DBLP:journals/corr/abs-1811-00855} and TAGNN \cite{10.1145/3397271.3401319}, running an extensive grid search on their parameters and using the best configurations found on train/validation split from \cite{ludewig2019performance}. We confirm the correctness of our configurations by cross-checking against results reported in the original author papers, using their original data splits (e.g. we achieve 0.1811 MRR@20 for TAGNN on DIGI dataset with the original author train/validation split, while the metric reported by authors is 0.1803 MRR@20). We confirm that the best configurations are the same for both original author data preprocessing and preprocessing from \cite{ludewig2019performance}. Following \cite{ludewig2019performance} we skip datasets requiring more than a week to complete hyperparameter tuning. Therefore we mark TAGNN on RSC15, 30MU, NOWP and SR-GNN on 30MU, RSC15 as timeout.

\textbf{Metrics.} Hit rate (HR@20) and mean reciprocal rank (MRR@20) are based on measuring to what extent an algorithm is able to predict the immediate next item in a session. Precision (Precision@20), recall (Recall@20) and mean average precision (MAP@20) consider all items that appear in the currently hidden part of the session.

\textbf{EMDE Configuration.} 
Items interactions are grouped by user, artist (for music datasets) and session. Then we create a graph with those interactions, where items are graph nodes with edges between them if an interaction appears. We embed these graphs with \cite{cleora}. As a result, we obtain an embedding vector for each item. For every dataset we perform a small random search over  number of iterations $[1,2,3,4,5]$, embedding dimension $[512, 1024, 2048, 4096]$, items interactions $[user, session, artist]$ and different sources of multimodal data depending on the dataset. We also observe that adding random sketch codes (not based on LSH) for each item improves the model performance, allowing the model to separate very similar items to differentiate their popularity.
All input modalities are presented in Table \ref{session-rec-input} in the Appendix.

We train a three layer residual feed forward neural network with 3000 neurons in each layer, with leaky ReLU activations and batch normalization.
We use Adam \cite{DBLP:journals/corr/KingmaB14} for optimization with a first momentum coefficient of 0.9 and a second momentum coefficient of 0.999  \footnote{Standard configuration recommended by \cite{DBLP:journals/corr/KingmaB14}}.
The initial learning rate was decayed by parameter $\gamma$ after every epoch. Refer to the Appendix for our full hyperparameter configuration.

The input of the network consists of two width-wise L2-nor\-ma\-li\-zed, concatenated, separate \textit{sketches}, one of them represents the last single item in the session, the second one all other items. In order to create representation of user's behaviour for the second \textit{sketch}, we aggregate the sketches of user's items, multiplying them with exponential time decay. Decay of sketch between time $t_1$ and $t_2$ is defined as:
\begin{equation}
sketch({t_2}) = \alpha w^{(t_2-t_1)}sketch({t_1})
\end{equation}
The parameters $\alpha$ and $w$ define decay force.

We perform a small random search over network parameters for each dataset, which is summarized in Table \ref{session-rec-conf} in the Appendix. For all datasets we used sketch of width $K = 7$ and $w=0.01$.

\textbf{EMDE Performance.} As presentend in Table \ref{session-rec-results}, EMDE outperforms competing approaches for most metric-dataset combinations, although \cite{ludewig2019performance} find that neural methods generally underperform compared to simple heuristics. The results with multimodal data shown in Table \ref{session-rec-mm-results} (\texttt{EMDE MM}) are significantly better than the basic configuration in all cases. Thus, the ability to easily ingest multimodal data proves important in practical applications.

\subsubsection{Top-k Recommendation}\label{k_top_experiment}

We conduct experiments on two popular, real-world, large-scale datasets: Netflix Prize \cite{Bennett07thenetflix} and MovieLens20M. The datasets contain movie ratings from viewers. We reuse the code for dataset preprocessing, training and evaluation from  \cite{ludewig2019performance}\footnote{\url{https://github.com/MaurizioFD/RecSys2019\_DeepLearning\_Evaluation}}.

\textbf{Benchmarks.} We compare against the baselines used by \cite{10.1145/3298689.3347058}, including a recent state-of-the-art VAE-based neural model: MultVAE \cite{10.1145/3178876.3186150}, and a non-neural algorithm EASE \cite{10.1145/3308558.3313710}.

\textbf{EMDE Configuration.} On top of EMDE, we use a simple one-hidden-layer feed-forward neural network with 12,000 neurons.  We put 80\% of randomly shuffled user items into the input sketch, and the remaining 20\% into the output sketch to reflect train/test split ratio of items for a single user. In our multimodal configuration we include the interactions of users with disliked items (items which received a rating lower than 4). Detailed information on item embeddings configuration can be found in Table \ref{top-k-input} in the Appendix.

\textbf{EMDE Performance.} The results are show in Table \ref{top-k-results}. We observe that our approach consistently and significantly outperforms the baselines for lower \textit{k} values in the top-k recommended item rankings for Movielens20M, which is consistent with CMS being a heavy-hitters estimator. In practice, the very top recommended items are key for user satisfaction as they are given the most attention by users, considering the limitation of item display capabilities and user's attention in the real world. For Netflix, we are able to outperform all competitors only on top-1 recommendation, which is probably caused by  comparatively lower density estimation scores we  achieve on this dataset (see Figure \ref{f:de_results}). This is probably due to the simplistic graph embedding method we use. Table \ref{tab:top-k-mm-results} presents results with added multimodal data (\texttt{EMDE MM}), which again are significantly higher than the unimodal case and competitors.

\subsection{EMDE as a Density Estimator}
\label{sec:density-estimator}
Most kernel density estimation methods such as \cite{10.5555/3037581.3037588} are not applicable for our solution since they do not scale to high input dimensionalities or require sampling during inference, thus preventing usage as input/output for neural networks. We compare against a recent SOTA fast hashing-based estimator (HBE) method for multidimensional data \cite{NIPS2019_9709}, which utilizes sampling during inference. This system can handle input dimensions on the scale of thousands, which is optimal for embedding systems currently in use. We repeat the Laplacian kernel testing procedure from \cite{NIPS2019_9709} on multiple datasets:  MovieLens20M and Netflix (embedded with \cite{cleora}), GloVe \cite{pennington2014glove} and MNIST \cite{lecun-mnisthandwrittendigit-2010}.

\begin{table}[t]
  \caption{Density estimation results. Metric reported is Pearson correlation coefficient against true distribution created as in \cite{NIPS2019_9709}.}
  \label{de_results}
  \centering
  \small
  \setlength\tabcolsep{2.5pt}
  \begin{tabular}{r|cccc}
  \hline
%   \toprule
  Method & GloVE & MNIST & MovieLens20M & Netflix \\\hline
  EMDE & 0.996 & 0.809 & 0.983 & 0.700 \\
  FastHBE & 0.997 & 0.998 & 0.986 & 0.695 \\
    \hline
  \end{tabular}
\end{table}

Results in Table \ref{de_results} show that with optimal parameter values EMDE is a competitive density estimator despite its simplicity. Pearson correlation is used as EMDE estimates are un-normalized. Figure \ref{f:de_results} shows the relation between N/K parameters and estimation quality. Large values of N have an ensembling effect due to averaging of smaller regions, increasing N is always beneficial. With too small K the number of created regions is low and the details in data distribution are not represented. Too large K makes similar data points spread over many regions instead of being contained in one representative region and the gain from the metric prior is lost. The optimal value of K needs to be established empirically.

Note that Fast HBE receives an advantage since the input space is constructed with Laplacian kernel, for which their method is purposefully aligned while EMDE is kernel-agnostic. In spite of this, EMDE still proves competitive.

\begin{figure}
\subfigure[Pure EMDE]{
  \includegraphics[width=0.15\textwidth]{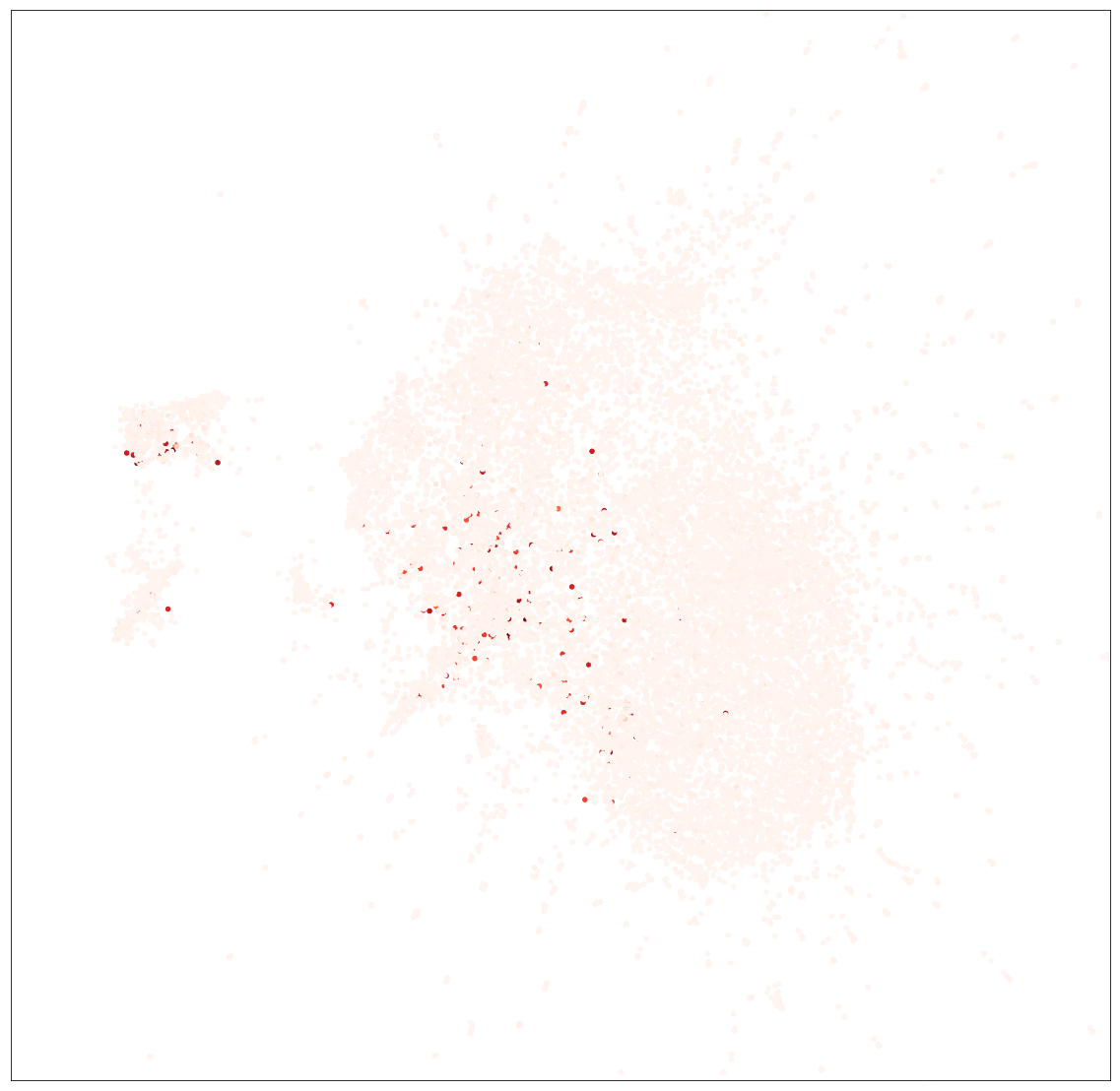}%
}
\subfigure[Cond EMDE]{
 \includegraphics[width=0.15\textwidth]{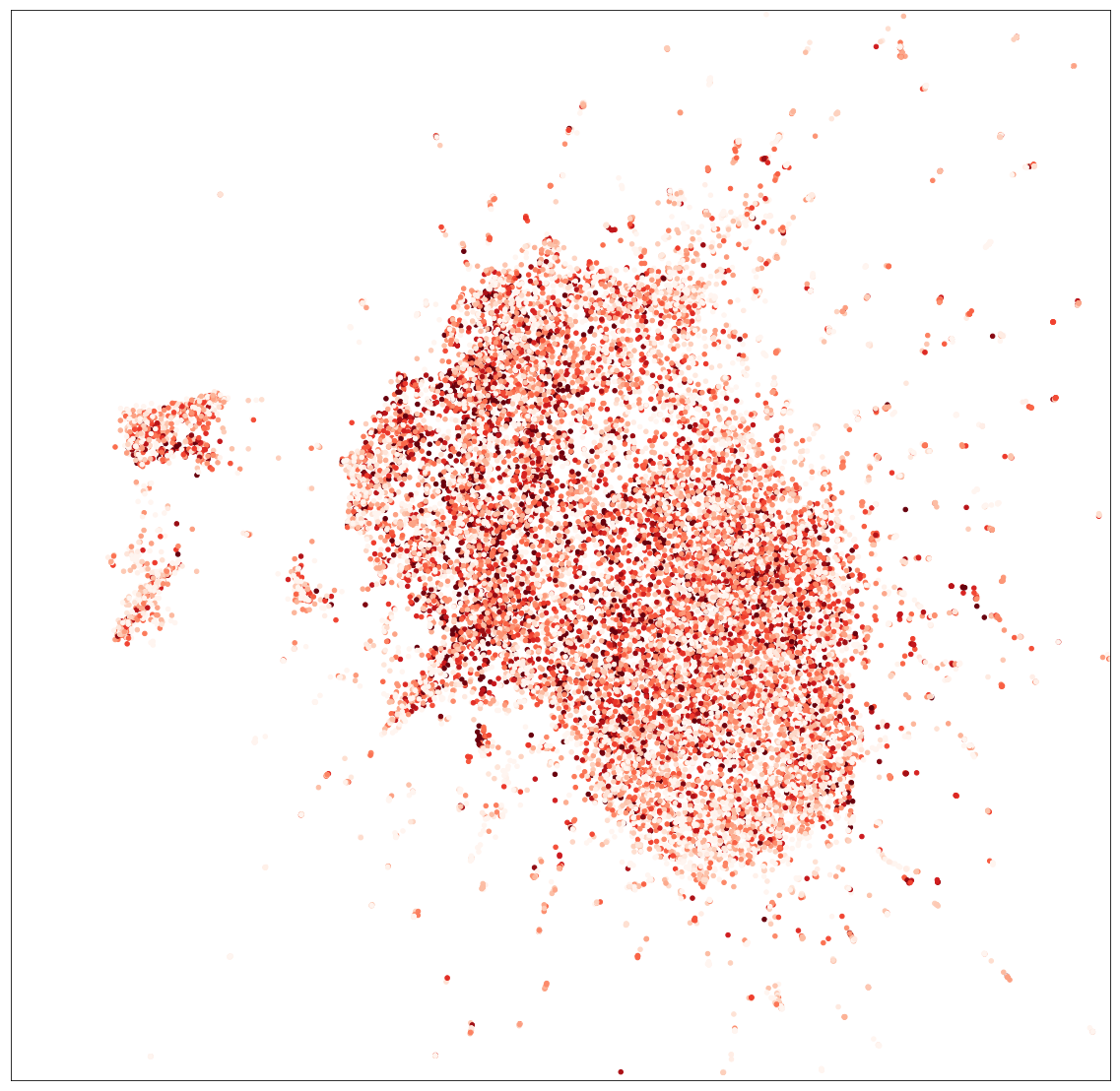}%
}
\subfigure[Item Popularity]{
  \includegraphics[width=0.15\textwidth]{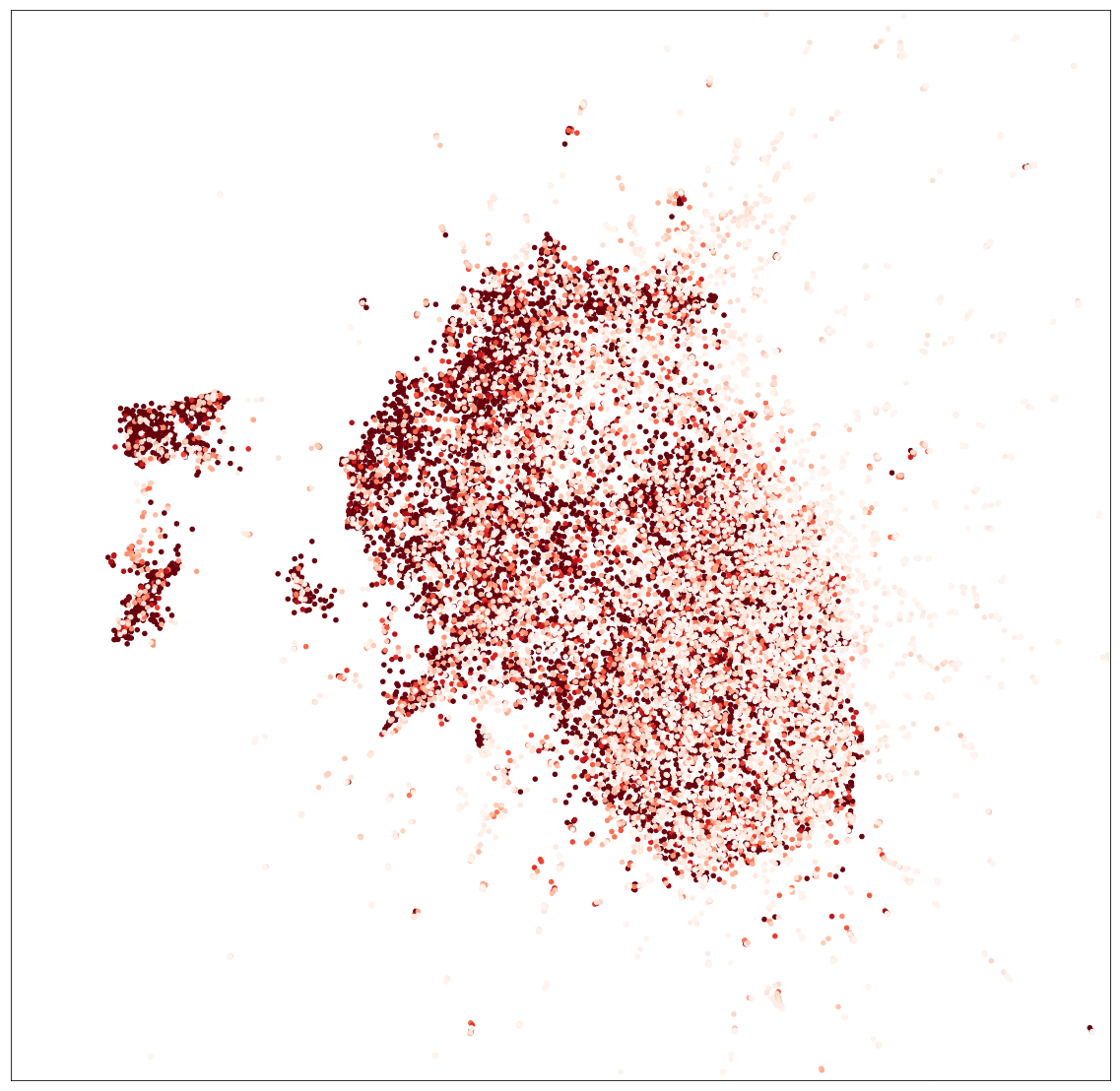}%
}
\caption{Example of the effects of 2 modes of operation of EMDE. \textbf{Pure EMDE} is a simple query on user liked items sketch. \textbf{Cond EMDE} has a neural network mapping from user liked item sketches to output sketches. We draw the plots by casting our embedded MovieLens item vectors onto a 2D plane with UMAP \cite{mcinnes2018umap-software} and scaling the color by the score a particular item gets with each mode of recommendation. Both plots are drawn for a single user. Darker color means higher score. \textbf{Item Popularity} shows item popularity scores across the whole dataset.}
\label{vis_conditional}
\end{figure}

\subsection{Ablation studies}\label{ablations}

\subsubsection{Pure vs. Conditional EMDE}
We study the benefits of using a neural network on top of EMDE. EMDE can be used in two ways:
\begin{enumerate}
    \item \textit{Pure EMDE}. The rationale behind this simple KNN-like approach is that users might like the items which are closest to their previous selection on the embedding manifold (located within the same or close buckets). It is done with a simple EMDE query on the user's liked items sketch.
    \item \textit{Conditional EMDE}. This approach involves a neural model for conditional estimation, which learns to map between input and output sketches. Reflecting complex patterns in user behavior, the model transforms the input sketch and produces a new sketch at output, on which EMDE query is performed. We use this approach in all our SRS and top-k experiments.
\end{enumerate}

\begin{table*}[!ht]
  \caption{Ablation study results. Bolded headers indicate parameters used in experiments.}
  \label{ablation results}
  \centering
  \small
  \setlength\tabcolsep{2.5pt}
  \begin{tabular}{r|lll|llll|llll}
  \hline
  \multicolumn{4}{c}{ Partitioning method} &
  \multicolumn{4}{c}{ Ensembling method} &
  \multicolumn{4}{c}{ Input embedding}\\
%   \toprule
  Metrics & \textbf{DLSH} & OPQ & PQ  & \textbf{gmean} & min & mean & hmean & 
  \textbf{MM} & interact & metadata & random \\
    \hline
    P@20 & 0.05564 & 0.05708 & 0.0519  & 0.05564 & 0.04634 & 0.05254 & 0.05172 & 0.0556 & 0.05067 & 0.05144 & 0.03044\\
    \hline
    
    MRR@20 & 0.36493 & 0.35937 & 0.35138 & 0.36493 & 0.31195 & 0.35922 & 0.34112  &  0.3649 & 0.35019 & 0.32813 & 0.27865\\
    \hline
    
    MAP@20 & 0.03016 & 0.03092 & 0.02838 & 0.03016 & 0.02508 & 0.02865 & 0.02785  & 0.0302 & 0.02747 & 0.02789 & 0.01685\\
    \hline
    
    R@20 & 0.49743 & 0.50752 & 0.47328 & 0.49743 & 0.42569 & 0.4786 & 0.46592  & 0.49743 & 0.46087 & 0.46769 & 0.30577\\
    \hline
    
    HR@20 & 0.62015 & 0.63284 & 0.58562 &  0.62015 & 0.52502  & 0.59597 & 0.57769 & 0.6202 & 0.57501 & 0.58113 & 0.37567 \\
    \hline
  \end{tabular}
  \label{ablation}
\end{table*}

\begin{table*}[t]
  \caption{Pure EMDE vs Conditional EMDE results on MovieLens dataset. Neural network setting is the same as in the experiment in top-k setting. \textbf{TopPop} is a baseline of most popular items.}
  \label{conditional results}
  \centering
  \small
  \setlength\tabcolsep{2.5pt}
  \begin{tabular}{r|cccc}
  \hline
  
%   \toprule
  Metric & Pure EMDE & TopPop & Pure EMDE+Pop & Cond EMDE \\\hline
  NDCG@20 & 0.0930 & 0.1201 & 0.1582 & 0.3053 \\
  Recall@20 & 0.1253 & 0.1441 & 0.2003 & 0.3523 \\
    \hline
  \end{tabular}
\end{table*}

A marked difference in preformance between these approaches on MovieLens20M  dataset is displayed in Table \ref{conditional results} (including also a version of pure EMDE with scores weighted by item popularity via simple multiplication). Conditional EMDE achieves a significant boost in scores due to a number of factors: 1) it learns a global structure, enabling it to model relationships between far away regions of the manifold, while pure EMDE models only the local structure (as evidenced by the spatial spread of high score points in Figure \ref{vis_conditional}); 2) it considers popularity of items where pure EMDE does not, yet only selectively (high score points in Figure \ref{vis_conditional} loosely correspond to clusters of popular items); 3) conditional EMDE is fully aligned with the training objective. In this way, conditional EMDE can hit the right balance between uniqueness and popularity, and the spatial distances between recommended items, while being able to model an arbitrary transformation of one density into another.

\subsubsection{Configuration Selection}

 In order to understand the effects of crucial parameters for training and decoding of EMDE, we conduct additional experiments in session-based scenario. We run experiments on RETAIL dataset, treating the best configuration for this dataset as our baseline. Results are reported in Table \ref{ablation results}.

\textbf{Manifold partitioning.} In addition to DLSH we verify the impact of partitioning the manifold with product quantization methods (PQ) which decompose high-dimensional space into the Cartesian product of low-di\-men\-sional subspaces which are quantized separately. We also analyze its enhanced and optimized version (OPQ) \cite{quantization}. We can see that DLSH is a strong baseline, leading at MRR@20 in comparison to other coders. However, we note that OPQ achieves competitive results, which indicates potential for improvement in density-based manifold partitioning methods.

\textbf{Score ensembling.} We perform point estimates from the output sketch by ensembling independent probability mass functions across sketch depth using geometric mean. While arguments behind the choice of geometric mean for ensembling probability measures can be found in \cite{dognin2019wasserstein} and \cite{itoh2017geometric}, we empirically confirm the choice, comparing with: minimum, arithmetic mean and harmonic mean.

\textbf{Choice of manifold} We investigate how the choice of input embedding manifold influences performance of downstream recommendation tasks. We compare our multi-modal item sketches (\texttt{MM}) with uni-modal sketches built on item-user interaction embeddings (\texttt{inter}), item attribute metadata embeddings (\texttt{metadata}), and random item embeddings without a metric prior (\texttt{random}). All sketches have the same dimensions for a fair comparison. Not only random item vectors have the lowest performance due to the lack of a metric prior, but the manifolds for item interaction similarity and item attribute similarity confer different benefits, outperforming each other in MRR@20 and P@20 respectively. The multi-modal sketch is a clear winner in both metrics.

\section{Conclusions}
We present EMDE - a compact density estimator for high dimensional manifolds inspired by CMS and LSH techniques, especially amenable to treatment with neural networks optimized with KL-divergence loss. We show that both sequential and top-k recommendation problems can be cast in a simple framework of conditional density estimation. Despite simplified treatment of sequential data as weighted multi-sets, our method achieves state-of-the-art results in recommendation systems setting. Easy incorporation of multiple data modalities, combined with scalablility and the potential for incremental operation on data streams make EMDE especially suitable for industry applications. Natural extensions to sequence-aware settings, such as item pair encoding or joint encoding of item and positional vectors are promising avenues for future research.

\bibliographystyle{unsrt}
\bibliography{sample-base}

\newpage

\appendix

\section{Experimental details}
Here we give an overview of experimental details for the results presented in the paper.  

\subsection{Session-based Recommendation: Details}

Our full hyperparameter configurations for each dataset (both basic \texttt{EMDE} and multimodal \texttt{EMDE MM}) are given in Table \ref{session-rec-conf}. Input embeddings for both \texttt{EMDE} and \texttt{EMDE MM} are described in Table \ref{session-rec-input}. 

\begin{table}
\caption{Input modalities of items for each top-k dataset.}
  \label{top-k-input}
  \centering
  \setlength\tabcolsep{2pt}
  \small
 
\begin{tabular}{l|l}
Dataset & input modalities \\ \hline
Movielens20M  & \begin{tabular}[c]{@{}l@{}}liked items dim 1024 iter 1\\ disliked items dim 1024 iter 1\\\end{tabular}                   
\\ \hline
\begin{tabular}[c]{@{}l@{}}Netflix \end{tabular} & \begin{tabular}[c]{@{}l@{}}liked items dim 1024 iter {[}1,2{]} \\ disliked items dim 1024 iter 1\\\end{tabular} 
\\ \hline
\end{tabular}
\end{table}

\begin{table}
\caption{Input modalities of items for each session-based dataset.}
  \label{session-rec-input}
  \centering
  \setlength\tabcolsep{2pt}
  \tiny
  \scalebox{0.8}{
\begin{tabular}{l|l||l|l}
Dataset & input modalities                                                                                                      & Dataset  & input modalities                                                                   \\ \hline
RETAIL  & \begin{tabular}[c]{@{}l@{}}session dim 4096 iter {[}2,3,4,5{]}\\ user dim 4096 iter {[}2,3,4,5{]}\\ random codes\end{tabular}                                      & \begin{tabular}[c]{@{}l@{}}RETAIL\\ MD\end{tabular} & \begin{tabular}[c]{@{}l@{}}session dim 1024 iter {[}2,4{]} \\ user dim 1024 iter {[}3{]}\\ property 6 iter 4 dim 1024\\ property 839 iter 4 dim 1024\\ property 776 iter 4 dim 1024\\ random codes\end{tabular} \\ \hline
DIGI    & \begin{tabular}[c]{@{}l@{}}session dim 1024 iter {[}1,2,5{]}\\ session dim 2048 iter {[}1,2,4,5{]}\\ random codes\end{tabular}                                     & DIGI MD                                             & \begin{tabular}[c]{@{}l@{}}session dim 1024 iter {[}1,2,3,4,5{]}\\ purchases iter 4 dim 1024\\ product names iter 4 dim 1024\\ item in search queries iter 4 dim 1024\\ random codes\end{tabular}               \\ \hline
RSC15   & \begin{tabular}[c]{@{}l@{}}session dim 1024 iter {[}2,3,4{]}\\ session dim 2048 iter {[}2,3,4{]}\\ random codes\end{tabular}                                       & RSC15 MD                                            & \begin{tabular}[c]{@{}l@{}}session dim 1024 iter {[}2,3,4{]}\\ session dim 2048 iter {[}2,3,4{]}\\ purchases iter 4 dim 1024\\ random codes\end{tabular}                                                        \\ \hline
30M & \begin{tabular}[c]{@{}l@{}}session dim 2048 iter {[}1,2,3,5{]}\\ artist dim 2048 iter {[}2,3,4,5{]}\\ user dim 2048 iter {[}2,3,4{]}\\ random codes\end{tabular}   & 30M MD                                          & \begin{tabular}[c]{@{}l@{}}session dim  2048 iter {[}1,2,3,5{]}\\ artist dim 2048 iter {[}2,3,4,5{]}\\ user dim 2048 iter {[}2,3,4{]}\\ playlist iter 4 dim 1024\\ random codes\end{tabular}                    \\ \hline
NOWP    & \begin{tabular}[c]{@{}l@{}}session dim 2048 iter {[}1,2,3,4{]}\\ artist dim 2048 iter {[}1,2,4,5{]}\\ users dim 2048 iter {[}2,4, 5{]}\\ random codes\end{tabular} & AOTM                                                & \begin{tabular}[c]{@{}l@{}}session dim 2048 iter {[}1,2,3{]}\\ artist dim 2048 iter {[}1,4,5{]}\\ users dim 2048 iter {[}1,2,3{]}\\ random codes\end{tabular}                                                     
\end{tabular}}
\end{table}

\begin{table*}
\caption{Training hyperparameters for each SRS dataset.}
  \label{session-rec-conf}
  \centering
  \tiny
  \setlength\tabcolsep{5pt}
\begin{tabular}{|l|l|l|l|l|l|l|l|l|}
\hline
Dataset                                                 & RETAIL & RETAIL MD & \begin{tabular}[c]{@{}l@{}}DIGI /\\ DIGI MD\end{tabular} & \begin{tabular}[c]{@{}l@{}}RSC15 /\\ RSC15 MD\end{tabular} & NOWP  & 30M & 30M MD & AOTM   \\ \hline
Epochs                                                  & 5      & 5         & 5                                                        & 7                                                          & 5     & 50      & 50         & 9      \\ \hline
Batch size                                              & 256    & 256       & 512                                                      & 512                                                        & 256   & 512     & 512        & 256    \\ \hline
lr & 0.004  & 0.004     & 0.004                                                    & 0.0005                                                     & 0.001 & 0.0005  & 0.0005     & 0.0005 \\ \hline
$\gamma$                                                   & 0.5    & 0.5       & 0.5                                                      & 1.0                                                        & 0.75  & 1       & 1          & 0.9    \\ \hline
$N$                                             & 10     & 10        & 10                                                       & 10                                                         & 10    & 10      & 9          & 9     \\ \hline
$\alpha$                                                   & 0.95   & 0.9       & 0.97                                                     & 0.9                                                        & 0.9   & 0.9     & 0.9        & 0.9    \\ \hline
\end{tabular}
\end{table*}

\subsection{Top-k Recommender System}
We empirically verify that a wide and simple network is optimal for the top-k problem. We train a simple one-layer feed-forward neural network with 12,000 neurons and leaky ReLU activations. We use Adam \cite{DBLP:journals/corr/KingmaB14} for optimization with a learning rate of 0.001. Batch size for each experiment equals 256.  

Input sketches are counted on interactions in the same way as in session-based recommendations (Section 1.1). For Netflix and Movielens20M, we use interactions of users with liked items (rating equal or above 4) and disliked items (rating below 4). We embed interactions with our basic embedding scheme, noting that the best results are achived on just 1 iteration of the embedding algorithm. This suggests that the tested top-k datasets are fundamentally different than session-based datasets, where valuable embedding vectors were obtained with even 4 or 5 iterations. Higher density of top-k datasets makes \cite{cleora} degrade with more iterations, rendering all items' embedding vectors too similar to each other and losing the cluster structure. 

We use input sketches of size 30x350 ($N$ x $2^K$), and output sketches (counted on just liked item interactions) are of size 30x350. Sketch sizes are estimated on the validation set.

\textbf{Datasets.}  After preprocessing, we obtain 20,108 movies and 116,677/10,000/10,000 users in the train/valid/test sets in MovieLens20M. The lowest number of movie ratings for single user is 5 (due to preprocessing constraints), the median is 38 and maximum is 3,177. 

For Netflix, we obtain 17,768 movies and 383,435/40,000/40,000 users in the train/valid/test sets. The lowest number of movie ratings for single user is 5 (due to preprocessing constraints), the median is 60 and maximum is 12,206.  

\textbf{Metrics.} We focus on metrics reported by \cite{10.1145/3298689.3347058}: Recall@K and NDCG@K. Recall is the total percentage of correct items retrieved in ranked item list cut off at the threshold \textit{K}. NDCG is a measure which additionally considers the exact rankings of the retrieved items - the closer to the actual user ranking, the higher the measure. We rank algorithms by their performance on Recall@20 and NDCG@20, adhering to \cite{10.1145/3298689.3347058}.

\end{document}